\pdfoutput=1

\documentclass[11pt]{article}

\usepackage{ACL2023}

\usepackage{times}
\usepackage{latexsym}

\usepackage[T1]{fontenc}

\usepackage[utf8]{inputenc}

\usepackage{microtype}

\usepackage{inconsolata}
\usepackage{comment}
\usepackage{booktabs}

\usepackage{graphicx}

\title{Emotion and Sentiment Guided Paraphrasing}

\author{Justin J. Xie\thanks{\hspace{.1 cm} Work done as a research intern at Portland State University.}\\
  Westview High School  \\
  Portland, OR, USA  \\
  \texttt{justinjxie@gmail.com} \\\And
  Ameeta Agrawal \\
  Portland State University  \\
 Portland, OR, USA \\
  \texttt{ameeta@pdx.edu} \\}

\begin{document}
\maketitle
\begin{abstract}
Paraphrase generation, a.k.a. paraphrasing, is a common and important task in natural
language processing. Emotional paraphrasing, which changes the emotion embodied in a 
piece of text while preserving its meaning, has many potential applications, including
moderating online dialogues and preventing cyberbullying. We introduce a new task of 
fine-grained emotional paraphrasing along emotion gradients, that is, altering the 
emotional intensities of the paraphrases in fine-grained settings following smooth 
variations in affective dimensions while preserving the meaning of the original text. 
We reconstruct several widely used paraphrasing datasets by augmenting the input and 
target texts with their fine-grained emotion labels. Then, we propose a framework for 
emotion and sentiment guided paraphrasing by leveraging pre-trained language models for 
conditioned text generation. Extensive evaluation of the fine-tuned models suggests that 
including fine-grained emotion labels in the paraphrase task significantly improves the 
likelihood of obtaining high-quality paraphrases that reflect the desired emotions while 
achieving  consistently better scores in paraphrase metrics such as BLEU, ROUGE, and METEOR.  

\end{abstract}

\section{Introduction}
With the rise of social media and online chat rooms, the textual aspect of language 
is often found to be the only aspect of communication that is transferred over the Internet. 
Devoid of any intonations or accompanying facial movements, it is more challenging for 
people to decipher the true meaning and underlying emotion that a message is intended to 
convey, especially if that message incorporates the more complex aspects of speech.
This could lead to negative social consequences. For example, political tweets from prominent figures without careful consideration can lead to political radicalization and conflicts. Furthermore, on messaging apps such as Discord, cyberbullies attack others with emotion-ladened words while innocent people send unnecessarily emotional messages in the heat of the moment. Emotional paraphrasing could be an important solution to overly intense emotions expressed on social media~\cite{empathicparaphrasing} and provide support toward moderation of hate speech \cite{tontodimamma2021thirty,enas23}.

Paraphrase generation (a.k.a. paraphrasing), a key task in natural language processing, involves generating an output text that preserves the meanings of the input text while including variations in words and grammars. 
The refined task of emotional paraphrasing  has garnered much recent attention \cite{casas2021emotional}. 
Its goal is to alter the underlying emotion associated with a sentence while maintaining its meaning. 

\begin{table*}[htb]
\centering
\begin{tabular}{ l | p{2.05cm} p{5.7cm} p{5.5cm}}
\toprule
\textbf{Dataset} & \textbf{Transition} & \textbf{Input Text} & \textbf{Paraphrased Text}\\
\midrule
\texttt{Google} & anger $\rightarrow$ \newline disappointment & {\em He is angry to learn that in June Ethan Lovett (Nathan Parsons) is his half brother.} & {\em He is upset to learn in June that Nathan Parsons (Ethan Lovett) is his half brother.} \\

\midrule
\texttt{MRPC}   & approval $\rightarrow$ \newline realization & {\em The decision was among the most significant steps toward deregulation undertaken during the Bush administration.} & {\em The decision is among the far-reaching deregulatory actions made during the Bush administration.}	\\

\midrule
\texttt{Quora} &	fear $\rightarrow$ \newline nervousness	& {\em My boyfriend wants to kiss me and I kind of want to kiss him, but I've never kissed anyone and I'm scared I'll be terrible at it. What should I do?} & {\em My boyfriend is wanting to kiss me and I want to kiss him too, but I've never kissed anyone, and I'm nervous. What do I do?} \\	
    

\bottomrule
\end{tabular}\caption{Some sample instances of emotion paraphrasing from our reconstructed datasets.}
\label{DatasetEx} 
\end{table*}

In this paper, we introduce a new task of {\em fine-grained} emotional paraphrasing along emotion gradients, i.e., altering emotional intensities in fine grain following smooth variations in affective dimensions (e.g., from anger to annoyance) while preserving the overall meaning. First, we analyze and reconstruct existing paraphrasing datasets to adapt them for the current task. Next, we propose the concept of an emotion-transition graph where transitions are based 
on the fine-grained emotions and their emotion gradients as identified by  
GoEmotions~\citep{demszky2020goemotions}, and are constrained by specific goals of emotion transition. Then, we develop a framework for emotion and sentiment guided paraphrasing by leveraging several pretrained language models for conditioned text generation under zero-shot, few-shot, and fully supervised settings. Lastly, we conduct a comprehensive evaluation of the proposed framework with several datasets using metrics pertaining to both paraphrasing and emotion transition.

In all settings, the fully supervised and few-shot fine-tuned models showed significant 
improvements over the zero-shot base models, i.e., doubling the number of exact matches 
of desired fine-grained emotions while achieving consistently better scores in paraphrase metrics 
such as BLEU, ROUGE, and METEOR. Few-shot learning delivered competitive performances in all 
categories compared with fully-supervised. This study indicates that our fine-grained emotional paraphrasing framework has potentials in applications to specific scenarios, e.g., chat rooms, forums, and 
public online spaces. 

Specifically, our contributions include:
\begin{itemize}
    \item Reconstructed Emotion Paraphrase Datasets: Given existing paraphrase datasets, we apply a fine-grained emotion classification model to label the input text and target text of each paraphrase pair with their emotions (see examples in Table~\ref{DatasetEx}). A similar procedure is also applied to label each paraphrase pair with their sentiment intensities: neutral, low, or high. 
    \item Emotional Paraphrasing Models: Leveraging pre-trained language models, we propose a paraphrasing framework guided by emotion and sentiment transitions. 
    \item Evaluation: We conduct an extensive set of experiments to verify the effectiveness of the proposed approach.
\end{itemize}

\section{Related Work} 
This section discusses two main threads of related work: emotion classification and paraphrasing. 

\subsection{Emotion Psychology and Classification}
Emotions are a key component of human psychology, playing a role in many cognitive processes including learning, memory, decision making, and interpersonal communication \citep{oatley1994experience, tyng2017influences}.
Equally important is the role that emotions play in human-to-human interactions. 
Words can trigger emotional responses, both negative and positive. 
Without facial expressions, vocal intonations, or hand gestures, it is harder to communicate one’s emotions online. The intensities of words can be higher than what someone wants them to communicate. For example, 
someone could want to communicate frustration, but instead could come off as furious. Rooted in the psychology of communication and emotion, the need for lowering intensity of online communications inspires the task of fine-grained emotional paraphrasing.

In 1890, \citeauthor{james1890principles}
proposed fear, grief, love, and rage as a set of the most basic emotions. Then, \citet{plutchik1980general} introduced eight categories of emotions, which was followed by  \citet{ekman1992argument} who introduced his famous set of six basic emotions: fear, anger, joy, sadness, disgust, and surprise. These taxonomies form the basis of many early NLP experiments pertaining to emotions \cite{ mohammad2010emotions, 6511907}. Another classification produced by \citet{lazarus1994passion} included a list of 15 emotions. Recently a study done by \citet{cowen2017self} expanded on these classifications. By having human test subjects report on the emotions they felt while viewing videos, the study found that there were 27 emotion categories, in addition to a neutral emotion. 
This study also grouped these emotions into ``clusters.'' \citet{demszky2020goemotions} produced a similar set of 28 emotions that was used in the GoEmotions project. 
This project provided a labeled dataset of 58K texts and a model based on BERT~\citep{devlin2018bert} capable of classifying inputs into one of the 28 emotions. 
In addition, 
the GoEmotions project provided a heatmap showing the adjacency between emotions by continuous gradients as well as including a stratification of the emotions into groups (see Appendix~\ref{sec:appendix}). While the proposed approach can adopt any emotion taxonomy, our work follows the GoEmotions groups as guidance for structuring the proposed emotion transition graph. 

\subsection{Paraphrasing} 
Paraphrasing involves changing the wording of an input text while preserving its original meaning. 
Several stuides combine deep generative models with other modeling and training techniques:  
e.g., variations using reinforcement learning \citep{li2017paraphrase}, long short-term memory or LSTM \citep{gupta2018deep}, and stacked residual LSTM \citep{prakash2016neural}. 
Transformer-based text-to-text models such as BART \citep{lewis2019bart} and T5 \citep{raffel2020exploring} have become more popular for paraphrasing. 
Several studies have been conducted to improve these models' paraphrasing performance through combining Transformers and sequence-to-sequence models \citep{egonmwan2019transformer} and joint paraphrase learning \citep{min2020advancing}.

Emotional paraphrasing, a task that alters the underlying emotion associated with the input sentence while maintaining its meaning, has been closely studied. 
\citet{casas2021emotional} fine-tuned six GPT models (one for each emotion) for emotional paraphrasing, where the input text was paraphrased to fit one of Ekman's six emotional categories. 
Our new task, instead, stipulates a more fine-grained emotion categorization and paraphrasing. Our fine-tuned language models conduct emotional transitions based on the emotion of the input text, and is capable of transitioning to various emotions along emotion gradients on a transition graph.

Our task is also related to emotion or sentiment text style transfer. 
\citet{sundararaman2020unsupervised} proposed an unsupervised aspect-level approach to sentiment controllable style transfer. Other studies include a delete-retrieve-generate approach \cite{li2018delete} and a mask-infill approach \cite{wu2019mask} to sentiment style transfer. Through masked language modeling and transfer learning, \citet{mohammadibaghmolaei2023tet} adapted style transfer to transform texts into one of four emotions: anger, fear, sadness, and joy. While these tasks transfer text following certain emotion or sentiment styles, our task focuses on more flexible fine-grained emotion and sentiment transitions.   
 
As our task lowers emotion intensity of input texts, thereby lowering the strong psychological effects that intense 
emotional interactions can bring, it also relates to the task 
of positive reframing \citep{ziems2022inducing}. Both focus on 
altering the emotions of texts, while preserving its 
underlying connotations. However, the task of positive 
reframing emphasizes altering the input text into a positive 
emotion while our task does not transit every emotion into 
a positive one, but rather lowers the intensities of emotions, 
which allows negative and positive emotions alike. Our goal of lowering the intensity of emotion in text is 
related to, but different from the task of neutralizing bias \citep{pryzant2020automatically}. Neutralizing bias strives to eliminate all bias, which results in most paraphrased texts being classified as \textit{neutral}. Our task aims to preserve the base meaning and tone while lowering the {\em intensity} of the emotion in the input text. Thus, the paraphrase still expresses its original view or belief,  but in a less provocative or intense manner.


\section{Fine-Grained Emotional Paraphrasing}

\subsection{Problem Description}

Given an input text $t_i$ with emotion $e_i$ where $e_i$ belongs to an emotion adjacency group $\mathcal{E}$: $e_i \in \mathcal{E}$, the task of fine-grained emotional paraphrasing along emotion gradients is to paraphrase $t_i$ into $t_f$ where the emotion of $t_f$ is $e_f$ and $(e_f \in \mathcal{E}) \cap (e_i != e_f)$. Further constraints help to guide the emotion transitions along a specific affective dimension, e.g., lowering the sentiment intensity. If the intensity of $e_i$ is $s_i$ and that of $e_f$ is $s_f$, the refined condition is $(e_f \in \mathcal{E}) \cap (e_i != e_f) \cap (s_f < s_i)$. 

\begin{figure}[t]
    \centering
    \includegraphics[width=\columnwidth]{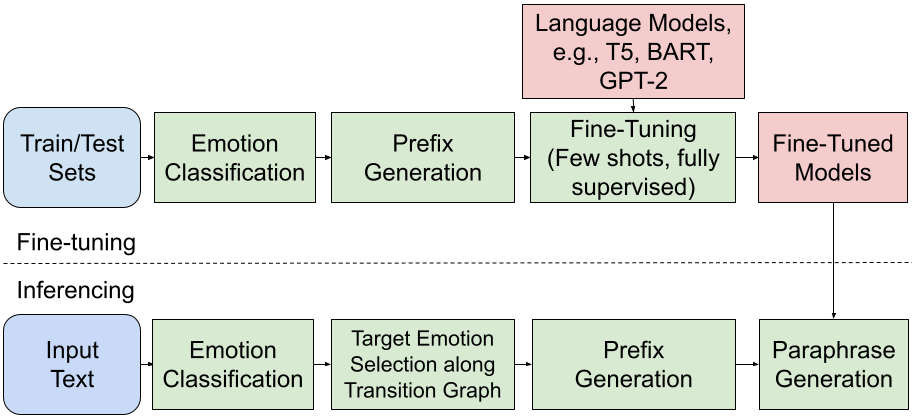}
    \caption{Workflow of Fine-Grained Emotional Paraphrasing along Emotion Gradients} 
    \label{fig:Approach1}
\end{figure}

To tackle the task of fine-grained emotional paraphrasing along emotion gradients, we propose a novel framework as illustrated in Figure~\ref{fig:Approach1}. 
The top part of this workflow fine-tunes pre-trained language models into fine-grained emotional paraphrasers. First, it labels the emotions of input and target texts of each paraphrase pair in both train and test sets. Then for each pair, a prefix of the form "(input emotion) to (target emotion)" is generated. Finally, the train/test sets augmented with emotion transition prefixes are utilized to fine-tune language models, e.g., T5, BART, and GPT-2, under three settings: zero-shot, few-shot, and fully supervised. The bottom of this workflow utilizes the fine-tuned paraphrasing models in inference applications. Given an input text $t_i$, it first identifies the emotion $e_i$ of $t_i$. Then it selects a target emotion $e_f$ for paraphrasing, utilizing an emotion transition graph that is based on emotion gradients. After that, it generates a prefix for the selected emotion transition "$e_i$ to $e_f$". Finally, it sends the query, "$e_i$ to $e_f$: $t_i$" to our fine-tuned paraphraser to generate the target paraphrase $t_f$.

\subsection{Emotion Classification}
The first step in our workflow is to identify the emotion ($e_i$) of the input text ($t_i$).
This is done through our enhanced version of the GoEmotions model: we modified the model to only report the dominant emotion that 
is above a certain threshold. If no emotion meets the threshold, the model reports no emotion label. 
Given the input text $t_i$, this classification model identifies the most compatible of the 28 emotions ($e_f$) to feed into the transition graph. 
The GoEmotions model has a wider variety and more detailed array of emotions compared to emotion classifications such as Ekman's. This allows for more precise emotion classifications that enable 
fine-grained adjustment of paraphrase emotions. 

\subsection{Target Emotion Selection Using Emotion Transition Graph}    
\begin{figure}[t]
\centering\includegraphics[width=\columnwidth]{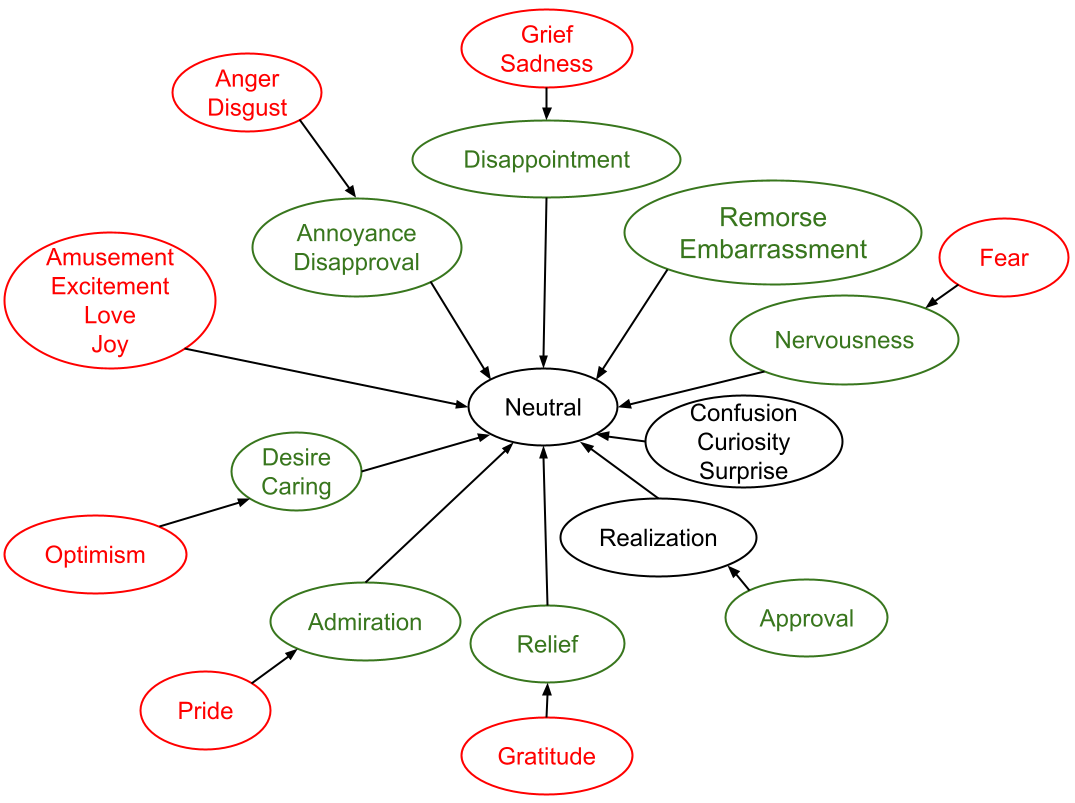}
    \caption{Sentiment Intensity Lowering Emotion Transition Graph: From High (Red) to Low (Green) to Neutral}
    \label{fig:TransitionGraph}
\end{figure}

The second step in our workflow is target emotion selection using an emotion transition graph such as the one shown in Figure~\ref{fig:TransitionGraph}. 
This particular transition graph is intended for lowering sentiment intensity. It is based on the GoEmotions emotion heatmap created by \citeauthor{demszky2020goemotions}, which shows emotions as grouped by continuous gradients. 
Each group of emotions (as shown in Appendix~\ref{sec:appendix}), 
although close in sentiments, exhibits different levels of intensities.
To measure the sentiment intensities of different emotions, we have applied NLTK's Vader Score~\citep{Hutto_Gilbert_2014} function to all emotion-labeled texts from the GoEmotions dataset and computed the median score for each emotion (which can be found in Appendix~\ref{sec:appendix}). Based on the median Vader scores of the 28 emotions, we are able to group them into five groups: high negative, low negative, neutral, low positive, and high positive as shown in Table~\ref{EmotionSentimentIntensityGroups}.   

\begin{table}[!t]
\centering
\noindent
\begin{tabular}{ cp{4.5cm}}
\toprule
\textbf{Group} & \textbf{Emotions} \\
\midrule
\texttt{high negative} & anger, disgust, grief, fear, sadness \\
\midrule
\texttt{low negative}  & nervousness, annoyance, disappointment, embarrassment, remorse, disapproval \\
\midrule
\texttt{neutral} & confusion, curiosity, realization, surprise, neutral \\
\midrule
\texttt{low positive} & approval, caring, desire, relief \\
\midrule
\texttt{high positive} & amusement, excitement, pride, optimism, gratitude, joy, admiration, love \\
              \bottomrule
\end{tabular}\caption{Emotion Grouping by Sentiment Intensity} 
\label{EmotionSentimentIntensityGroups} 
\end{table}
The emotion transition graph in Figure~\ref{fig:TransitionGraph} is derived by combining the two groupings found in GoEmotions and Table~\ref{EmotionSentimentIntensityGroups}. The emotions in red are emotions of high sentiment intensities, positive or negative, those in green are of low sentiment intensities, and those in black have neutral sentiment intensities. The arrows between ovals indicate the emotions in these ovals belong to the same GoEmotions emotion clusters, i.e., they are adjacent and connected with continuous gradients. The arrows to the neutral oval indicate that all emotions can transit to the neutral emotion.  
By following the transition graph, we can adjust emotion intensity.
For example, if the GoEmotions model identifies the input emotion as ``anger,'' the transition graph may recommend a transition to ``annoyance.'' 

\subsection{Prefix Generation}
The third step in our workflow is prefix generation. We adopt the multi-task design for text-to-text generation, i.e., many NLP tasks can be cast as text-to-text tasks and a prefix can be added to the input text to indicate the task at hand. Our prefix generator utilizes this design and generates the prefix for the task of fined-grained emotional paraphrasing. Given the source emotion $e_i$ identified in the emotion classification step and the target emotion $e_f$ selected in the target emotion selection step, the prefix is generated in the format of ``$e_i$ to $e_f$'' and placed in front of the input text $t_i$. It guides the fine-tuned language models to paraphrase along the selected emotion transition. An example of such a prefix would be: \textit{``anger to disappointment: He is angry to learn that in June Ethan Lovett (Nathan Parsons) is his half brother.''}

In addition, we also explore the use of sentiment ranges (i.e., high positive, low positive, neutral, low negative, and high negative) in place of fine-grained emotion labels as alternative fine-grained prefixes. Such a prefix would look like: \textit{``high\_neg to low\_neg: He is angry to learn that in June Ethan Lovett (Nathan Parsons) is his half brother.''}

\subsection{Paraphrase Generation}
The final step of our workflow is paraphrase generation which utilizes a fine-tuned language model to complete the task of fine-grained emotion paraphrasing along emotion gradients. Such a model is fine-tuned with a dataset of paraphrase pairs that exemplify the transitions along the continuous gradients that connect the emotions. The fine-tuned model allows for precise emotional paraphrasing by inputting the emotion transition prefix and the original text, paraphrasing it, and outputting the paraphrase that best fits the target emotion.

\section{Experiments}

Figure~\ref{fig:ExperimentWorkflow} illustrates the workflow of our experiments on preparing the train/test datasets for fine-grained emotional paraphrasing, conducting fine-tuning on various language models, and evaluating the emotional paraphrasing performance of these models. 
\begin{itemize}
\item Given a paraphrase dataset, we first label the input text and target text of each paraphrase pair with fine-grained emotions by using our modified verison of GoEmotions model. 
\item Second, we remove the paraphrase pairs that have the same input/target emotions and those pairs whose input or target emotions are labeled as neutral, as we are focused on the paraphraser's ability to lower the emotional intensity instead of neutralizing it. 
\item Third, we select the paraphrase pairs with decreasing intensity and if a pair has increasing intensity, we flip its input/target texts and emotions, so it can be used in our experiment. 
\begin{figure}[htb]
\centering\includegraphics[width=\columnwidth]{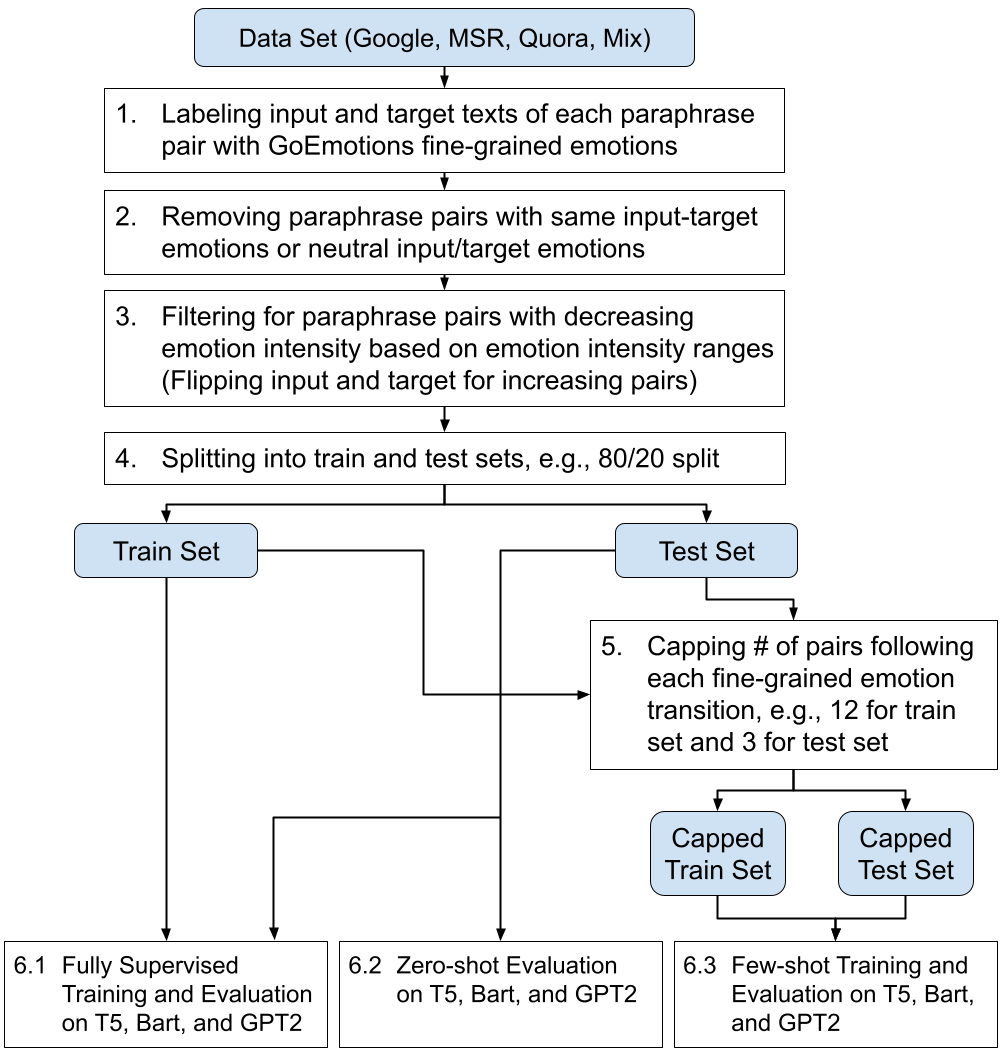}
    \caption{Experiment Workflow}
    \label{fig:ExperimentWorkflow}
\end{figure}
\item Fourth, we split the dataset into train/test sets, e.g., with a 80/20 split. 
\item Fifth, an optional step for few-shot training, we cap the number of instances of the same emotion transition, e.g., 12 in the train set and 3 in the test set following the 80/20 split. 
\item Sixth, we conduct three types of fine-tuning: fully supervised (or full), zero-shot, and few-shot and compare the performances of each type of fine-tuned model. In the zero-shot case, we directly evaluate the original model without fine-tuning and in the few-shot case, we fine-tune the model with the capped datasets as in Step 5 and evaluate with the full test set. 
\end{itemize}

\subsection{Datasets}\label{SubSec:DatasetPreparation}

Three publicly available paraphrasing datasets were used in our experiments after reconstruction. These include {\bf Google PAWS-Wiki} (PAWS) \citep{zhang2019paws}, {\bf Microsoft Research Paraphrase  Corpus} (MRPC) \citep{dolan2005automatically}, and {\bf Quora Questions Pairs} \citep{iyer2017qqpairs}. 
%

The Google PAWS project produced multiple sets of paraphrasing input-output pairs. 
We chose to use to PAWS-Wiki Labeled (Final) data because they were generated by translation methods and human verified for accuracy.
%
The MRPC corpus was a compilation of human-annotated data from the news. 
%
The Quora corpus has the goal of aiding the training of ``semantic equivalence'' models,
similar to the goals of paraphrasing models. Some sample instances are presented in Table~\ref{DatasetEx}. 


To make these datasets suitable for our emotional paraphrasing task, we reconstructed them by following Steps 1-4 in Figure~\ref{fig:ExperimentWorkflow}. 
The statistics of the filtered datasets are shown in Table~\ref{DatasetStats}, and these datasets are also combined into a Mix dataset for the study of overall performance. 

\begin{table}[!t]
\small
\centering
\def\arraystretch{1.25}
\setlength{\tabcolsep}{4.5pt}
\begin{tabular}{l|c|c|c|c}
\toprule
          &  {Total}  & {Emotion}    & {Emotion}   & {Sentiment}\\
{Dataset} &  { \# of}  & {Transiting}   & {Transiting}  & {Intensity}\\
          &  {Pairs}  & {w/ Neutral} & w/o Neutral & {Lowering}\\
\midrule
 PAWS  & 57401 & 3593 & 432   & 395 \\ 
\midrule
 MRPC    & 3728  & 508 & 53    & 32 \\ 
\midrule
 Quora   & 149263 & 32866 & 16935 & 2401 \\ 
 \midrule
 \midrule
 Mix     & 210392 & 36967 & 17420 & 2828 \\ 
 \bottomrule
\end{tabular}
\caption{Dataset Statistics} 
\label{DatasetStats} 
\vspace{-0.05in}
\end{table}


\begin{table*}[h!]
\centering
\begin{tabular}{ c | l | l | c c c c c}
\toprule
& & & \multicolumn{2}{c@{}}{\textbf{Emotion-Transition}}  & \multicolumn{3}{c@{}}{\textbf{Paraphrasing}} \\
& \textbf{Training} & \textbf{Prefix Type} & Exact-SR & Exact-FE & BLEU & R-L & METEOR \\
\midrule
 & Full & Sentiment Ranges & \textbf{0.796} & \textbf{0.632} & 0.314 & \underline{0.557} & 0.571 \\
 &  & Fine-grained Emotions & \textbf{0.801} & \textbf{0.604} & 0.316 & 0.555 & \underline{0.572} \\
T5 & Few-Shot & Sentiment Ranges & 0.791 & 0.620 & 0.298 & 0.528 & 0.547 \\
 &  & Fine-grained Emotions & 0.698 & 0.534 & 0.301 & \underline{0.538} & \underline{0.561} \\
 & Zero-Shot & Sentiment Ranges & 0.450 & 0.349 & 0.248 & 0.484 & \underline{0.515} \\
 &  & Fine-grained Emotions & 0.468 & 0.307 & 0.244 & \underline{0.488} & 0.513 \\
\midrule
 & Full & Sentiment Ranges & 0.719 & 0.606 & \textbf{0.408} & \textbf{\underline{0.626}} & \textbf{0.663} \\
 &  & Fine-grained Emotions & 0.706 & 0.578 & \textbf{0.409} & \textbf{0.619} & \textbf{\underline{0.665}} \\
BART & Few-Shot & Sentiment Ranges & 0.719 & 0.606 & 0.408 & \underline{0.626} & 0.663 \\
 &  & Fine-grained Emotions & 0.706 & 0.578 & 0.409 & 0.619 & \underline{0.665} \\
 & Zero-Shot & Sentiment Ranges & 0.291 & 0.339 & 0.335 & 0.588 & 0.633 \\
 &  & Fine-grained Emotions & 0.290 & 0.237 & 0.335 & 0.588 & 0.633 \\
 \midrule
 & Full & Sentiment Ranges & 0.691 & 0.494 & 0.168 & 0.381 & 0.399 \\
 &  & Fine-grained Emotions & 0.649 & 0.471 & 0.164 & \underline{0.387} & \underline{0.407} \\
GPT-2 & Few-Shot & Sentiment Ranges & 0.668 & 0.461 & 0.150 & 0.371 & 0.391 \\ 
 &  & Fine-grained Emotions & 0.639 & 0.452 & 0.178 & \underline{0.389} & \underline{0.408} \\
 & Zero-Shot & Sentiment Ranges & 0.632 & 0.113 & 0.004 & \underline{0.094} & \underline{0.124} \\
 &  & Fine-grained Emotions & 0.593 & 0.080 & 0.005 & 0.091 & 0.117 \\
\bottomrule
\end{tabular}\caption{Evaluations of T5, BART, and GPT-2 for Fine-Grained Emotional Paraphrasing}  
\label{Mix-Results} 
\end{table*}


\subsection{Evaluation Metrics}

The emotional paraphrasing capabilities of the models are evaluated from two aspects: \textbf{emotion transition} and \textbf{paraphrasing}. 

To evaluate the emotion transition performance of the models, we utilize the {\em Exact} metric to compute two scores: {\em Exact-SR} and {\em Exact-FE}. The {\em Exact-SR} score measures the percentage of the emotion sentiment ranges (i.e., high positive, low positive, neutral, low negative, and high negative) of the generated paraphrases that match the target sentiment ranges. The {\em Exact-FE} score measures the percentage of the fine-grained emotions of the generated paraphrases that match the target emotions. By comparing the sentiment ranges and specific emotions of the target texts and the predictions of each model, the {\em Exact} scores indicate how capable a model is at emotion transitioin. 
%

To evaluate the paraphrasing capabilities of the models, we utilize several metrics:
{\em BLEU} \cite{papineni2002bleu}, {\em ROUGE-L} \cite{lin2004rouge}, and {\em METEOR} \cite{banerjee2005meteor}. 
They evaluate the similarities of target texts and model predictions. 
%




\subsection{Models}
Below we discuss our models and training settings. 

\noindent \textbf{Emotion Labeling.} 
The original GoEmotions model, for each input text, outputs a list of emotions that it identified as being ``possible" candidates for the emotion of the input text and a confidence score for each candidate.  
In our experiments, we modified the model to only report the dominant emotion with a confidence score over 0.5. 



\noindent \textbf{Paraphrasing.} 
For paraphrasing, we fine-tuned 3 pre-trained language models, T5, BART, and GPT-2. 
We adopted multi-task training. 
Let $t_i$ be the input text and $e_i$ be its emotion. Let $e_f$ be the target emotion, and $t_f$ be the emotional paraphrased output of $t_i$. 
In the task of fine-grained emotional paraphrasing along emotion gradients, $t_i$, $e_i$, and $e_f$ are given to the language model in the query format: ``$e_i$ to $e_f$: $t_i$''. 
The fine-tuned model will output $t_f$,  a paraphrased version of $t_i$ where the underlying semantics of $t_i$ is kept and the intensity of emotion is changed. 
Each model is trained under 3 settings: fully supervised, few-shot, and zero-shot.

\subsection{Implementation}
We utilized the Simple Transformers package \citep{rajapakse2022simpletransformers} Version 0.63.6 to fine-tune T5 and BART models. 
For GPT-2, we utilized Huggingface's transformers implementation~\citep{huggingface:gpt2} Version 4.25.1. We conducted fine-tuning and evaluation on a desktop with an AMD Ryzen 7 5800x, 32GB RAM, and RTX 3080TI GPU. Due to a limited amount of GPU memory, 12GB precisely, we had to adopt a smaller batch size of 6. Each model was fine-tuned over 3 epochs. 



\section{Results and Discussions}

\begin{figure*}[htb]
\centering\includegraphics[width=0.85\textwidth]{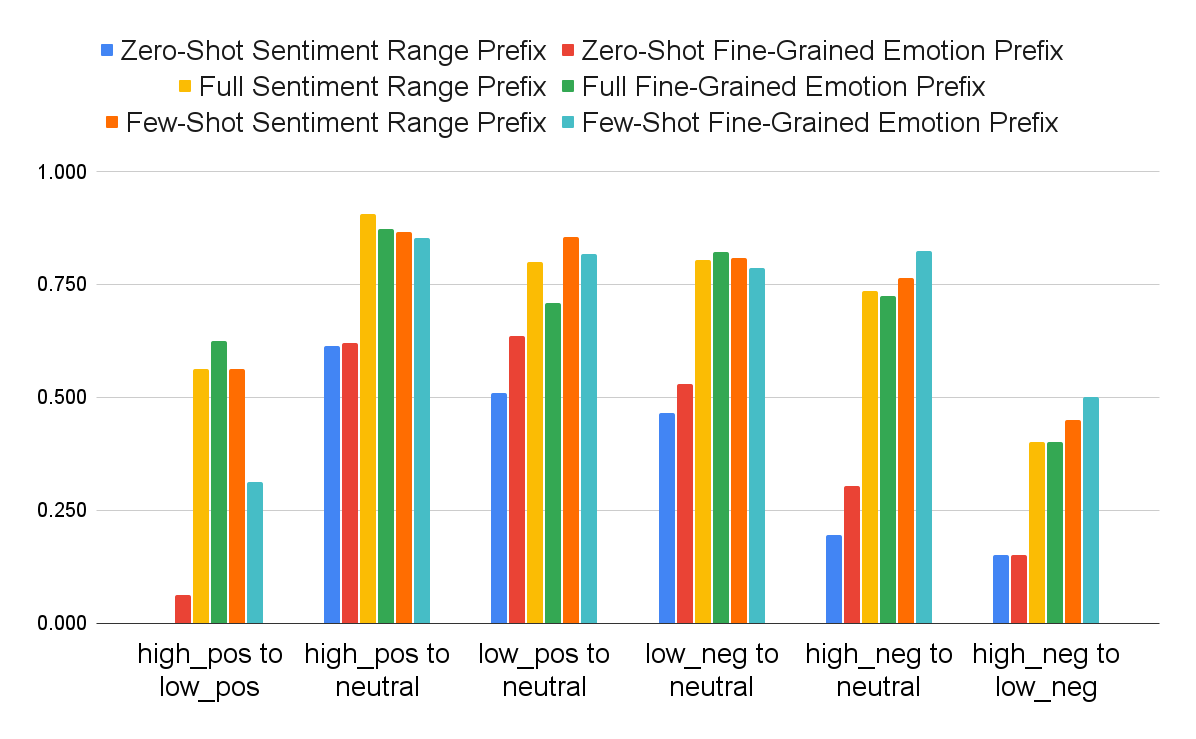}
    \caption{Success Rates of T5 in Transitioning Sentiment Intensity Levels on Mix Dataset} 
    \label{fig:Sen-Range-Trans-Rates}
\end{figure*}

\begin{figure*}[htb]
\centering\includegraphics[width=0.75\textwidth]{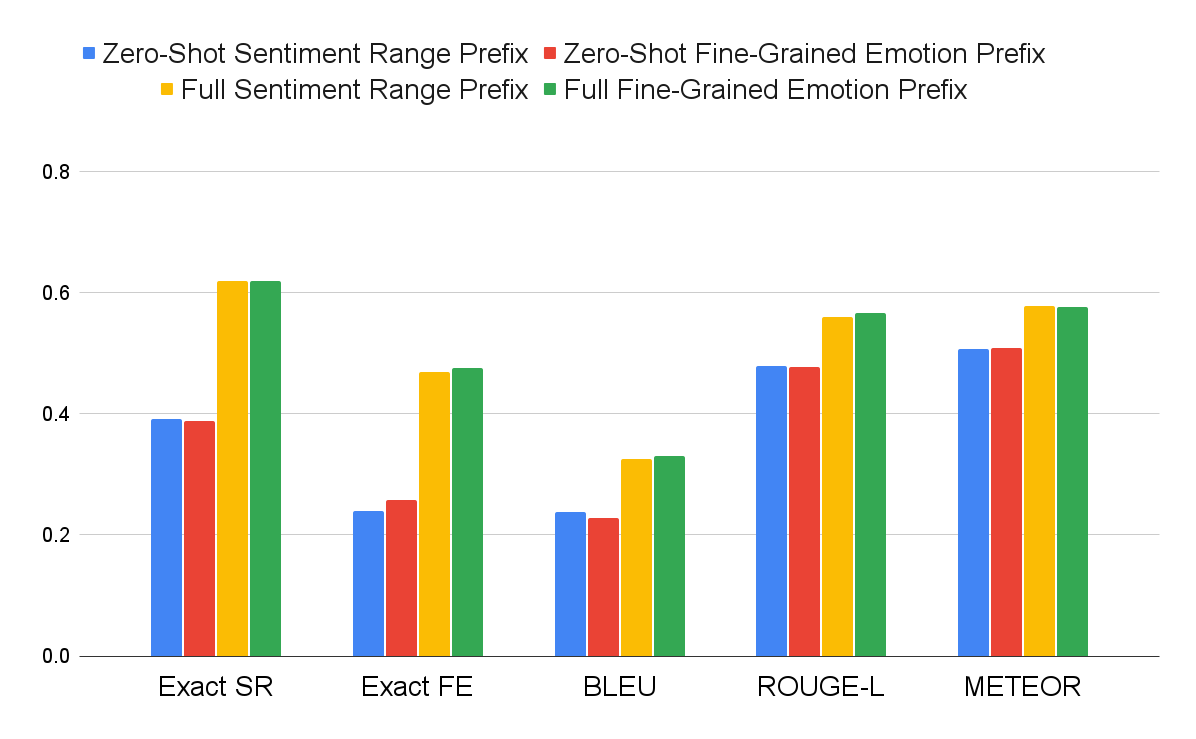}
    \caption{Fine-Tuned T5 Models on Test Dataset Enhanced by Transition-Graph-Guided Emotion Selection} 
    \label{fig:tg-enhanced}
\end{figure*}
  

Table~\ref{Mix-Results} summarizes the results from our experiments using T5, BART, and GPT-2 models for the fine-grained emotional paraphrasing task. It can be observed for all three models, fully supervised fine-tuning significantly outperformed the zero-shot setting in every category in both emotion-transition and paraphrasing metrics. Few-shot fine-tuning delivered competitive performances in all categories compared with the fully supervised setting.

When comparing model performance, it can be observed that T5 outperforms BART and GPT-2 on emotion-transition. This may be attributed to T5's design as a multi-task model meant to accept the prefixes we utilized. 
For paraphrasing, BART outclassed both T5 and GPT-2 models in text similarity and consistency. We speculate that designing more appropriate prompts might benefit GPT-2. 


In few-shot fine-tuning, we experimented with different limits for the numbers of text pairs following each fine-grained emotion transition in the train/test sets, 4/1, 8/2, 12/3, 16/4, and 20/5 per the 80/20 split. All few-shot train/test sets delivered better emotional transition performance than zero-shot and their paraphrasing performance became consistently better with 12/3 split and above. 

One important takeaway from the results is the similarity in performance of using sentiment ranges or fine-grained emotions as part of the prefix prompt to the models. We noticed that there was an insignificant difference in both the emotion-transition and paraphrasing performances of the two prefix types. 
An explanation for this behavior in the fine-tuned models may be that emotion transitions largely follow continuous gradients among emotions along certain affective dimensions and, therefore, lowering the sentiment intensity from an emotion often transitions to a same target emotion. 
This means that although the prefixes are different, the models learn the same emotion transitions that are embodied in the paraphrase pairs. 

Figure~\ref{fig:Sen-Range-Trans-Rates} illustrates the success rates of T5 in transitioning texts between different sentiment intensity levels under different fine-tuning settings. We observe that fully supervised and few-shot fine-tuning both outperform zero-shot significantly in all sentiment intensity lowering transitions. Fully supervised seems to perform better in emotion transitions lowering positive sentiments while few-shot better in lowering negative sentiments. Importantly, we also observe that lowering from high positive or negative to low positive or negative is more challenging for the model than lowering to neutral level.

\section{Case Study on Transition Graph Guided Target Emotion Selection} 

We created a new test dataset from the original Mix test dataset by leveraging transition-graph-guided emotion selection. Instead of utilizing the target emotion provided by the original test dataset, the transition-graph was used to randomly select a new target emotion that would maintain the emotion proximity while lowering the emotional intensity. However, if the neutral emotion was selected, the original target emotion was kept. In doing so, 35 percent of the dataset was given a larger variety of transition types between the high, low, and neutral emotion groups, while the size of the dataset was maintained. 
The emotion of the model prediction was compared to the desired target emotion to evaluate emotion-transition performance. The model prediction was compared to the original target text for measuring paraphrasing performance. 

Figure~\ref{fig:tg-enhanced} shows the performances of zero-shot and fully supervised 
fine-tuned T5 models on this new test dataset. They continue to reflect the observation from 
Table~\ref{Mix-Results} that the fine-tuned models show major improvements in emotion transition, 
while maintaining a slight gain in paraphrasing performance.  With the increased variety of target 
emotions, the success rate of the models does decrease as indicated by the 
lower {\em Exact} metrics. This points to the neccesity of paraphrase datasets that provide better 
coverage of the emotion transition graph which helps automate the target 
emotion selection for practical emotion moderation applications. 

\section{Conclusions and Future Work}

In this paper, we introduced a new task of fine-grained emotional 
paraphrasing along emotion gradients. 
We developed a workflow for addressing this task 
by fine-tuning pre-trained language models under multi-task 
learning framework. 
Our experiments have demonstrated that fine-tuned models 
perform significantly better than baseline models in 
both emotion transition and paraphrasing. 


For future work, there is still much to improve for fine-grained emotional paraphrasing. 
We will pursue better datasets for emotional fine-tuning or even develop 
new datasets for this purpose. We will further develop our approach on top 
of the state-of-the-art large language models, e.g., GPT-4. We will also
investigate more customized models beyond the baseline language models. 
For evaluation, we plan to conduct human studies as appropriate.




\section*{Limitations}
There is no dataset currently available specific for fine-grained emotional paraphrasing. 
For our study, we have to utilize publicly available paraphrase datasets, 
Google PAWS, MRPC, and Quora and augment their text pairs with emotions labels.
These datasets may not be best suited for studying this new task.   
Therefore, new datasets that are particularly developed for 
fine-grained emotional paraphrasing are needed. Furthermore, it is also
desirable to evaluate the proposed methods in alternative application scenarios other 
than lowering sentiment intensity. 

When using GoEmotions as our fine-grained emotion classifier, we selected the
emotion with the dominant confidence score above the threshold of 0.5. As the 
authors of GoEmotions have pointed out, there is still much room to improve 
on the classification accuracy of GoEmotions. Although the confidence score threshold 
of 0.5 worked well in our experiments, how to set this threshold still requires 
more studies. Similarly we utilized NLTK's Vader scores to place emotions into
high, low, and neutral intensity groups. The Vader score thresholds for this grouping 
were selected empirically. Further studies are needed for setting the
thresholds or developing better ways for intensity grouping. 

In the evaluation of our fine-grained emotional paraphrasing models, we utilized 
two sets of metrics for emotion transition and paraphrasing respectively. It is
desirable to jointly evaluate these two aspects, which we believe would be best 
done by well-designed human studies in future work.

\section*{Ethics Statement}
Our study is based on publicly available datasets from reputable 
sources. 
The augmented datasets will be made available with open-source code release. 
The fine-grained emotional paraphraser obtained through our study is based on
existing pre-trained language models and paraphrase datasets; therefore, it 
may inherit their drawbacks such as undesirable social biases. As an unintended 
use, the methods proposed by this paper can be utilized or modified to produce 
paraphrasers that increase the emotional intensities of texts, leading to 
texts with extreme emotions that can be potentially harmful. 
While we advocate 
for voluntary adoption of emotion moderation to achieve more peaceful cyberspaces, 
we do realize that the proposed methods can be abused as emotion moderation tools 
for censorship. We strongly oppose such applications.

\section*{Acknowledgments}
We thank the anonymous reviewers for their helpful feedback.

\bibliography{anthology,FGEPAEG}
\bibliographystyle{acl_natbib}

\vspace{0.025in}
\appendix

\section{Appendix}
\label{sec:appendix}

\begin{table}[htb]
\centering
\def\arraystretch{1.5}
\begin{tabular}{ c | l }
\toprule
\textbf{Group} & \textbf{Emotions} \\
\midrule
 1 & neutral \\
 2 & amusement, excitement, joy, love \\
 3 & optimism, desire, caring \\
 4 & pride, admiration \\
 5 & gratitude, relief \\
 6 & approval, realization \\
 7 & surprise, curiosity, confusion \\
 8 & fear, nervousness \\
 9 & remorse, embarrassment \\ 
 10 & disappointment, sadness, grief \\
 11 & disgust, anger, annoyance, disapproval\\
 \bottomrule
\end{tabular}
\caption{Emotion Grouping by \citet{demszky2020goemotions}}
\label{EmotionClusters} 
\end{table}

\begin{table}[htb]
\centering
\noindent
\def\arraystretch{1.5}
\begin{tabular}{ c | c }
\hline
\textbf{Emotions} & \textbf{Median Vader Score} \\
\hline
grief	& -0.5423 \\
anger	& -0.5234 \\
disgust	& -0.51805 \\
fear	& -0.4404 \\
sadness	& -0.4404 \\
\hline
nervousness	& -0.3597 \\
disappointment & -0.3059 \\
annoyance &	-0.296 \\
embarrassment &	-0.26655 \\
remorse	& -0.0772 \\
disapproval & -0.0644 \\
\hline
confusion &	0 \\
curiosity &	0 \\
realization & 0  \\
surprise & 0  \\
neutral	& 0 \\
\hline
approval & 0.296  \\
caring & 0.3412  \\
desire & 0.4019  \\
relief & 0.4391  \\
\hline
amusement &	0.4404  \\
excitement & 0.4404  \\
pride &	0.4767  \\
optimism &	0.5081  \\
gratitude &	0.5574  \\
joy	& 0.6008  \\
admiration & 0.6249  \\
love &	0.6369  \\
\end{tabular}
\caption{Sentiment Intensities of Emotions by NLTK Vader Scores Computed on GoEmotions Dataset} 
\label{EmotionMedianVaderScores} 
\end{table}

\begin{figure*}[htb]
\centering\includegraphics[width=0.73\textwidth]{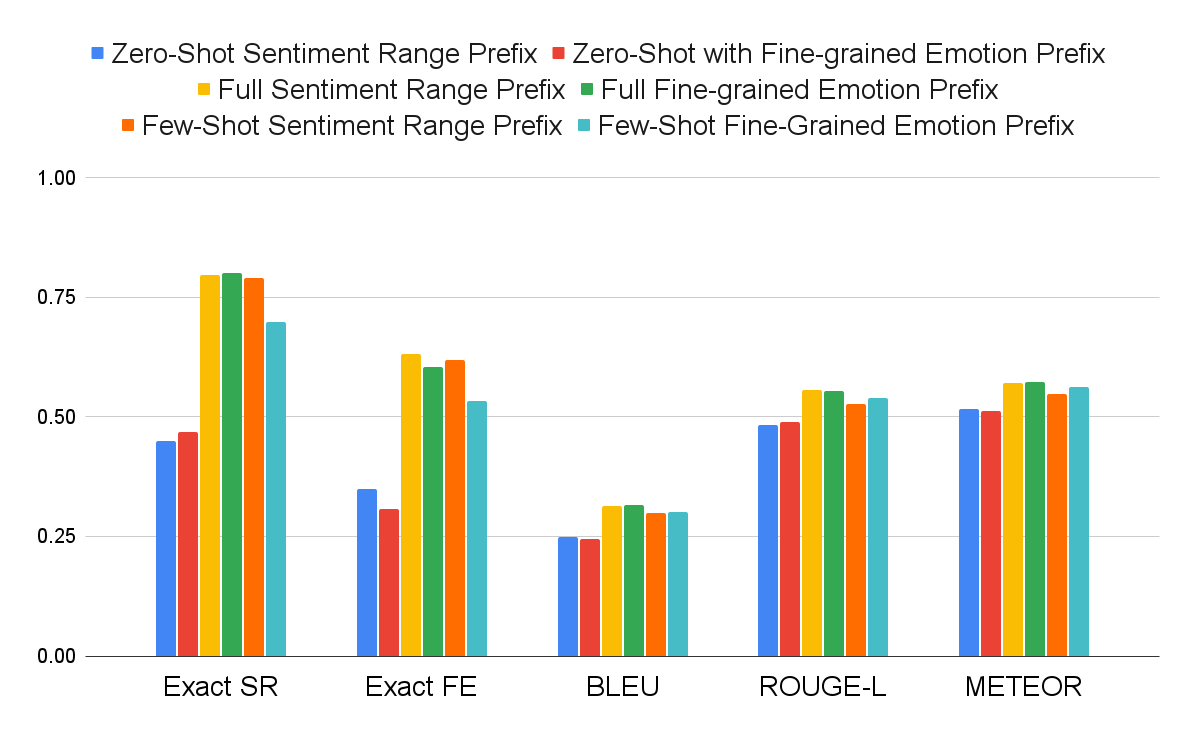}
    \caption{Evaluation Results of Mix Dataset on T5}
    \label{fig:T5-Mix-Results}
\end{figure*}

\begin{figure*}[htb]
\centering\includegraphics[width=0.73\textwidth]{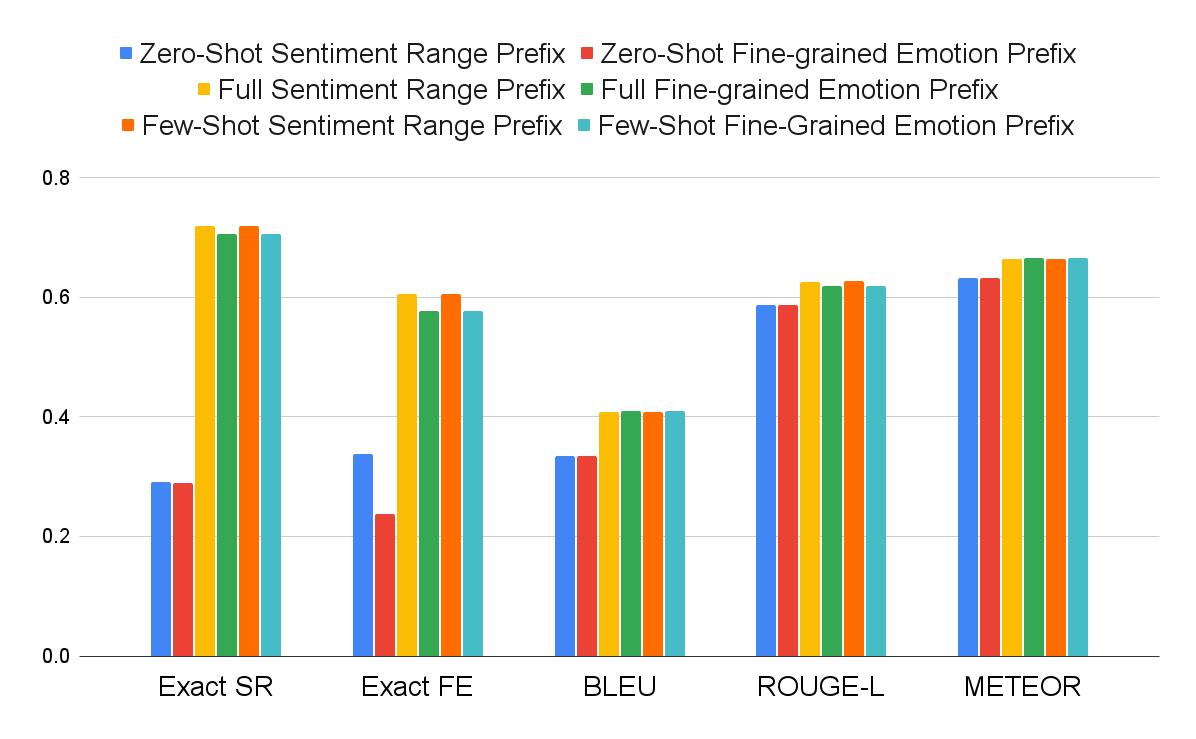}
    \caption{Evaluation Results of Mix Dataset on BART}
    \label{fig:Bart-Mix-Results}
\end{figure*}

\begin{figure*}[htb]
\centering\includegraphics[width=0.73\textwidth]{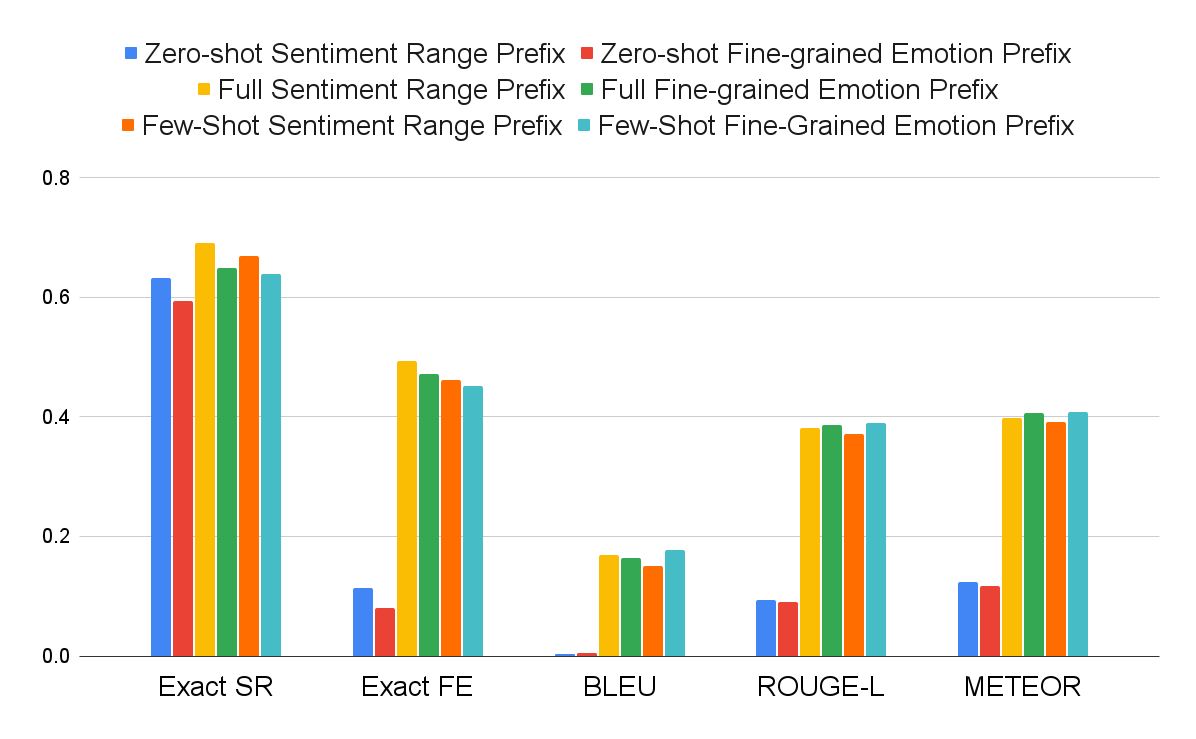}
    \caption{Evaluation Results of Mix Dataset on GPT2}
    \label{fig:GPT2-Mix-Results}
\end{figure*}

\begin{figure*}[htb]
\centering\includegraphics[width=0.73\textwidth]{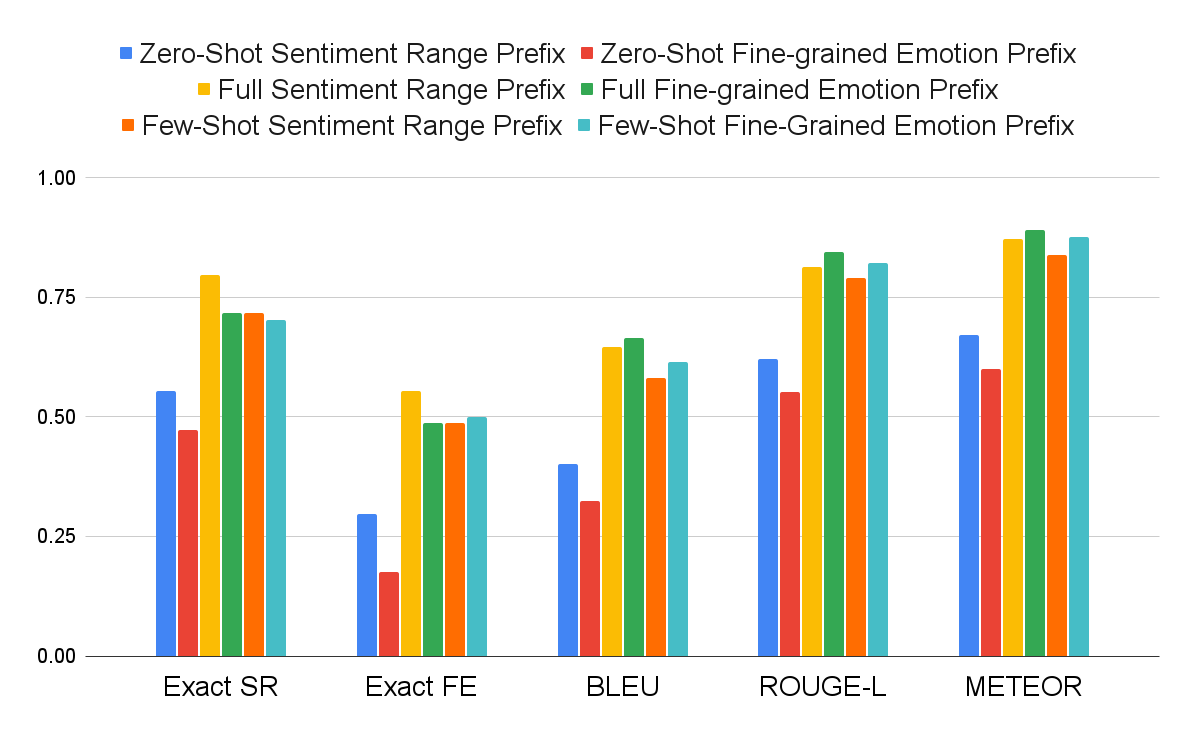}
    \caption{Evaluation Results of Google Dataset on T5}
    \label{fig:T5-Google-Results}
\end{figure*}

\begin{figure*}[h]
\centering\includegraphics[width=0.73\textwidth]{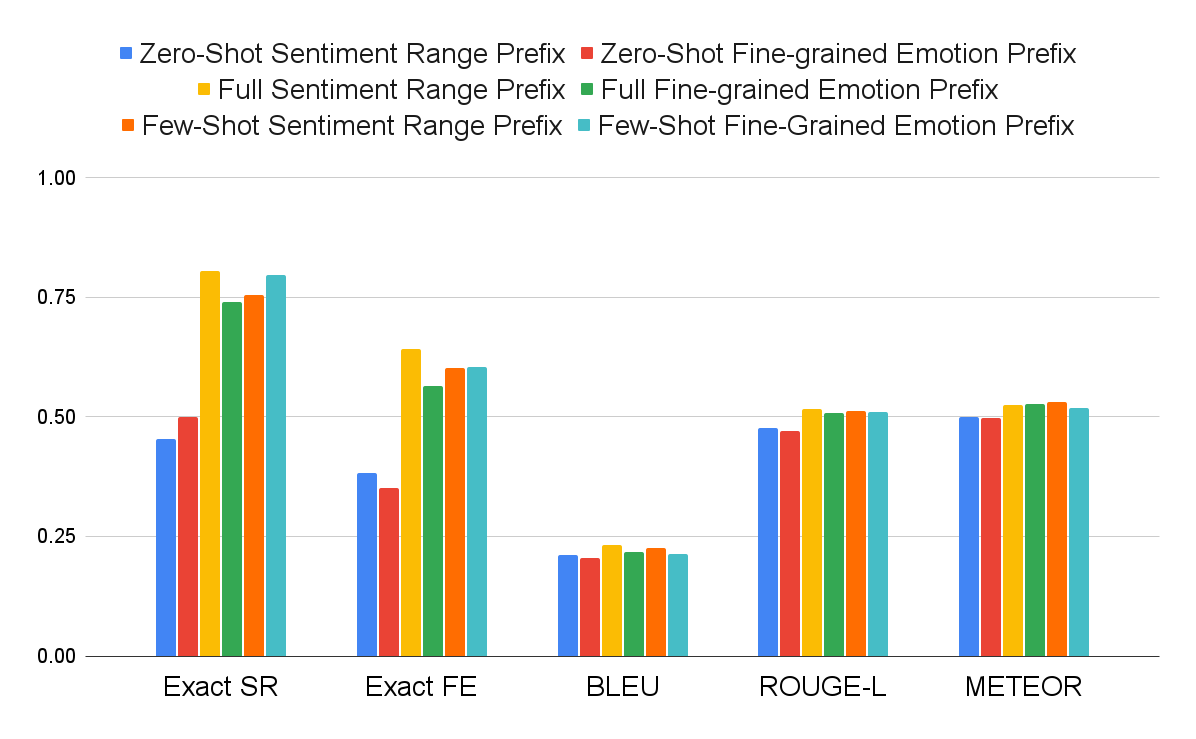}
    \caption{Evaluation Results of Quora Dataset on T5}
    \label{fig:T5-Quora-Results}
\end{figure*}


\end{document}